\documentclass[letterpaper, 10 pt, conference]{ieeeconf}  

\IEEEoverridecommandlockouts                              

\usepackage{amsmath} 
\usepackage{amssymb}  

\usepackage{graphicx,import}
\usepackage{subcaption}
\usepackage{makecell}
\usepackage{tabularx}
\usepackage{xcolor}
\usepackage{booktabs}
\usepackage{bm}
\usepackage{multirow}

\usepackage{pgfplots}
\pgfplotsset{compat=1.18}          
\usepackage{tikz}
\usetikzlibrary{plotmarks}
\usepackage{grffile}
\usepackage{amsmath}

\newlength{\figurewidth}
\newlength{\figureheight}

\title{\LARGE \bf
A Transparent 3-DoF Assistive Device for Upper-Limb Tremor and Ataxia Compensation: Design and Experimental Validation
}

\author{Thomas Dobosz*$^{1,2}$ and Veronika Hofmann*$^{1,2}$, Ahmed Barhoma$^{1}$, Ann-Kathrin Henke$^{1}$, Johannes A.~Gaus$^{3,6}$, \\Hans Mees$^{3}$, Winfried Ilg$^{3,6}$,
Isabell Wochner$^{3,6}$, Martin Giese$^{3,6}$, Andreas Bulling$^{4,6}$, Syn Schmitt$^{5,6}$,\\ Christophe Maufroy$^{1,2}$, Urs Schneider$^{1,2,6}$, Thomas Bauernhansl$^{1,6}$, and Daniel Haeufle$^{3,6}$
\thanks{*These authors contributed equally to this work.}
\thanks{
This work was financed by the Baden-Wurttemberg Stiftung in the scope of the AUTONOMOUS ROBOTICS project iAssistADL granted to WI, MG, SS, and DH. $^{1}$ Institute of Industrial Manufacturing and Management, University of Stuttgart, Germany; $^{2}$ Biomechatronic Systems, Fraunhofer IPA, Stuttgart, Germany; $^{3}$ Hertie Institute for Clinical Brain Research and Centre for Integrative Neuroscience, University of Tubingen, Germany; $^{4}$\,Institute for Visualization and Interactive Systems, University of Stuttgart, Germany; $^{5}$ Institute for Modelling and Simulation of Biomechanical Systems, University of Stuttgart, Germany; $^{6}$ Center for Bionic Intelligence Tubingen Stuttgart (BITS), Germany. The authors thank the International Max Planck Research School for Intelligent Systems (IMPRS-IS) for supporting Johannes A. Gaus.
}}
\begin{document}

\maketitle
\thispagestyle{empty}
\pagestyle{empty}

\begin{abstract}

Assistive devices for patients with neurodegenerative movement disorders must combine high mechanical transparency with the ability to generate corrective interaction forces. Achieving both properties simultaneously remains challenging. This paper presents the completed iAssistADL platform, a 3-DoF assistive device for future tremor and ataxia compensation during activities of daily living. The system employs a transparency-oriented mechanical architecture with all actuators located at the base, resulting in low moving mass, decoupled kinematics, and high mechanical transparency. To compensate for friction-induced transparency limitations along the vertical axis, a hybrid control strategy is proposed that combines open-loop force rendering in the horizontal plane with closed-loop force control in the vertical direction. The platform is experimentally evaluated through transparency measurements and static and dynamic force-rendering experiments at representative configurations. 
Results demonstrate resistive torques below 0.1\,Nm in the horizontal plane, end-effector forces of up to approximately 10\,N throughout the workspace, and accurate horizontal force rendering up to 4 Hz. While the vertical axis remains limited by friction-induced dynamics at higher frequencies, the proposed system successfully combines high transparency with controlled force generation, establishing a promising hardware platform for future tremor and ataxia compensation.


\end{abstract}




\section{Introduction}

Activities of daily living (ADLs) such as drinking or eating are essential for independent living and quality of life. In patients affected by neurodegenerative movement disorders, including Parkinson's disease or cerebellar ataxia, these activities are often impaired by pathological movement components such as tremor or dysmetric over- or undershooting. While voluntary movement generation is largely preserved, involuntary motions interfere with task execution \cite{Bhanpuri2014,Hermle2024, Oh2024, ParedesAcuna2024, Sun2023}. 

\begin{figure}[t] \centering \includegraphics[width=0.8\columnwidth]{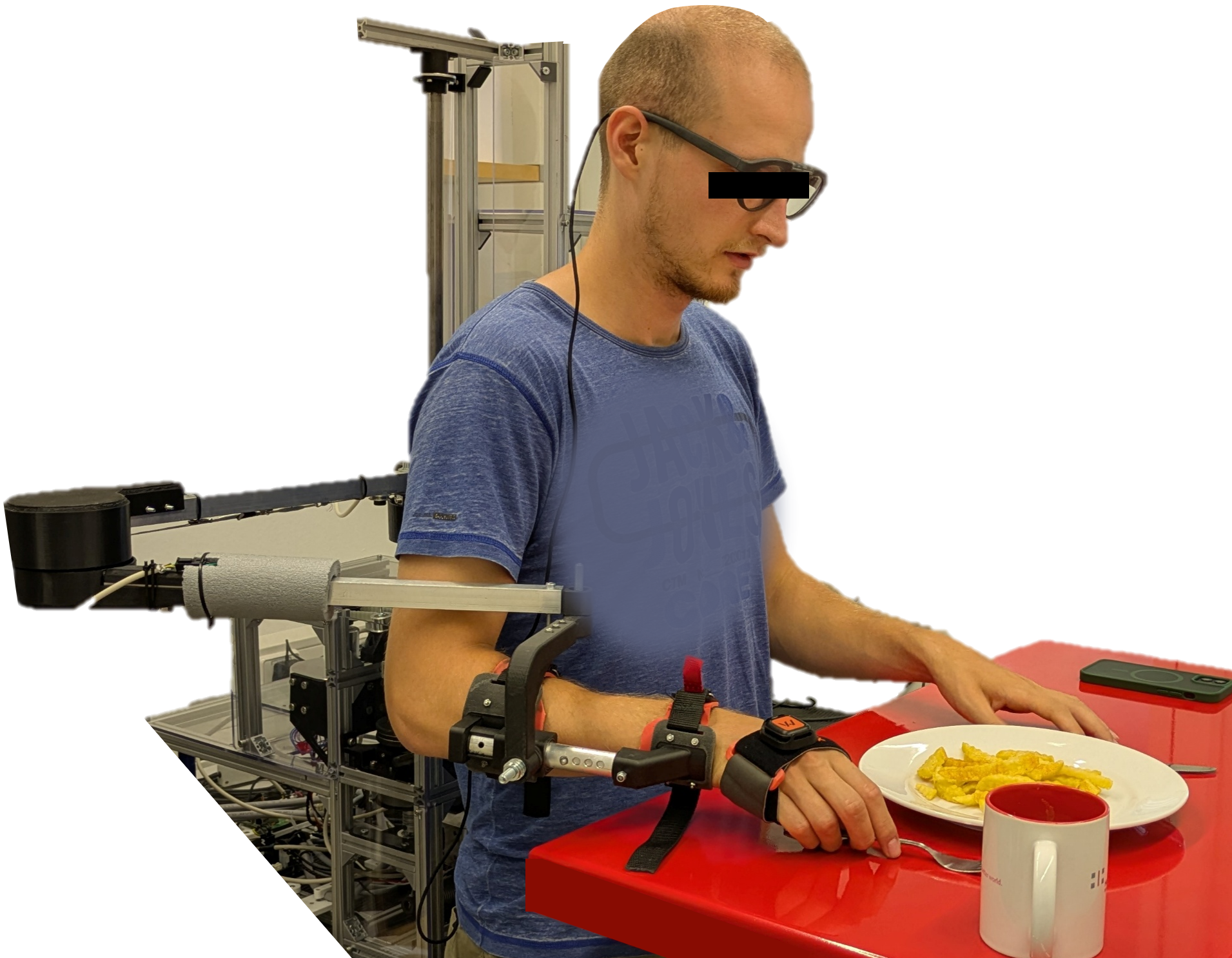} \caption{User interacting with the iAssistADL platform. The device applies assistive forces to the forearm while preserving natural upper-limb motion.}\label{fig:user} \end{figure}

In contrast to stroke rehabilitation devices, assistive technologies for this patient group should not replace movement generation but rather support the user by selectively compensating for pathological motion while preserving voluntary motor control \cite{LoraMillan2021}. Existing assistive devices often face a trade-off between transparency and force capability, as force-capable systems often introduce friction, inertia, and resistance that interfere with voluntary user motion. Achieving high transparency while maintaining sufficient force rendering capability remains a key challenge. Related end-effector robots such as MIT-MANUS and H-Man similarly address the trade-off between low mechanical impedance and active force rendering, but target rehabilitation and motor-control applications rather than compensation of pathological movement components \cite{Krebs2004,Campolo2014}.

Two families of devices exist for suppressing pathological upper-limb motion. Grounded systems are mounted to a table or wheelchair and slow the limb with velocity-dependent resistance \cite{Rosen1995, Aisen1993}. Wearable orthoses act at single joints instead, using dampers \cite{Loureiro2005}, driven transmission \cite{Rocon2007,ZHOU2017}, or electrical stimulation \cite{Dosen2015}.
In either case, whatever resists the tremor also resists the user. A systematic review of wearable orthoses found that some residual resistance always remains, whether the mechanism is passive, active, or semi-active. The best-performing device in the review also removes 17\% of the voluntary movement along with the tremor \cite{Fromme2019}. Users perceive this resistance and expend additional muscle effort to maintain the intended movement \cite{Verdel2024}.


Unlike wearable tremor-suppression devices, iAssistADL is intended as a stationary assistive platform mounted to a wheelchair or table environment, targeting upper-limb activities of daily living such as eating and drinking (see Fig.\,\ref{fig:user}). This platform is grounded like CEDO \cite{Rosen1995}, but it generates corrective forces instead of only removing energy. Damping can slow oscillation, but it has no preferred direction and therefore cannot correct a reach that overshoots its target. Dysmetric overshoot in cerebellar ataxia requires a force directed towards the intended endpoint, which only active force rendering can provide. The ataxia literature supports this: added limb mass improves some single-joint reaches but does not carry over to multi-joint movement \cite{Zimmet2019}, whereas forces applied through a laboratory exoskeleton and tailored to the individual patient did reduce reaching error \cite{Bhanpuri2014}.

This paper presents and experimentally validates a method for combining high mechanical transparency with effective force rendering in an assistive ADL platform. Building upon our previous work \cite{Ilg2025}, which introduced the system concept and passive transparency characterization, the present work adds the completed 3-DoF platform, the hybrid force-control strategy, and its experimental force-rendering validation.
The resulting system is experimentally evaluated through transparency measurements and force-rendering experiments at representative ADL configurations, assessing its suitability as a hardware platform for future compensation of pathological movement components, such as dysmetric overshoot or tremor.

\section{Methods}
\begin{figure}[t] \centering \includegraphics[width=\columnwidth]{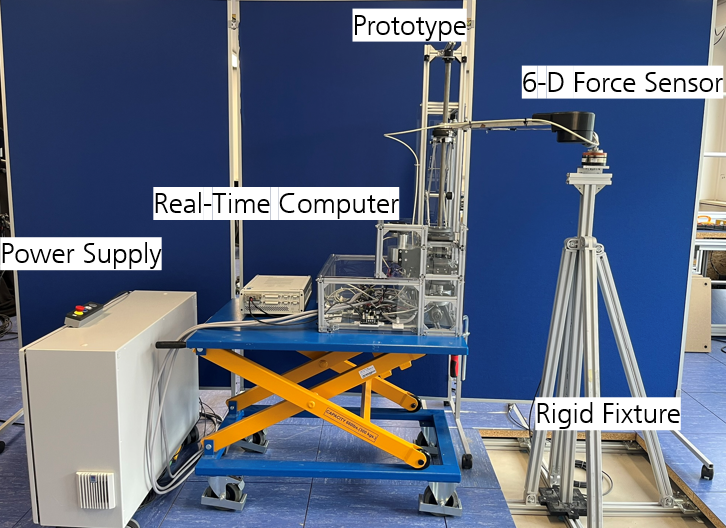} \caption{iAssist prototype along with the power supply and real-time computer. Prototype end-effector connected to a 6-D force sensor on a rigid fixture.} \label{fig:aufbau} \end{figure}

The mechanical architecture of the iAssistADL platform prioritizes transparency in the dominant horizontal movement directions, denoted as the x- and y-axes of the system.
Previous neuromechanical simulations within the iAssistADL project indicated that corrective forces of approximately 5\,N are sufficient to compensate relevant tremor and ataxic movement deviations during representative ADLs \cite{Ilg2025}. 
Since the assistive forces are applied at the forearm rather than directly at the hand, the device was designed to provide interaction forces of up to 10\,N to account for the more proximal application point and inter-subject variability. In a previous study, we experimentally characterized the passive transparency of each axis \cite{Ilg2025}. While the horizontal axes exhibited only minor resistive torques, substantial friction was observed along the vertical z-axis. To compensate for this behavior, a hybrid control strategy was developed. The resulting controller combines open-loop force generation in the horizontal plane with closed-loop force control in the vertical direction.
After characterizing the transparency of all three DoF, the control architecture was extended to actively render interaction forces. The force rendering capability was evaluated experimentally at three representative positions corresponding to grasping a bottle, pouring water, and drinking. Step and sinusoidal force references were applied and compared to measurements obtained from an external six-axis force sensor (ATI Axia80, Novanta Inc., USA) mounted between the end-effector and a rigid fixture (see Fig. \ref{fig:aufbau}).

\subsection{System Design}

\begin{figure} [thbp]
\centering 
\def\svgwidth{\columnwidth}
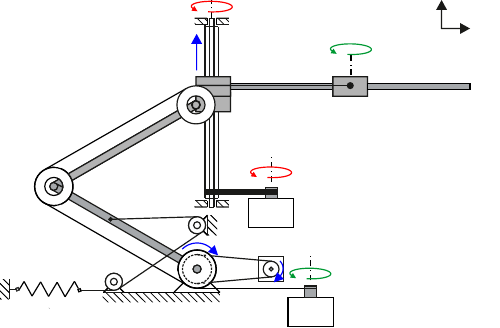
\caption{Schematic of the assistive device with three active DoF. A prismatic joint along the z-axis (blue) actuated via
a scissor mechanism, whose motion is constrained by the ball-spline
mechanism, a first rotational joint along the ball-spline axis (red), and a
second rotational joint bilaterally actuated using cables routed through
the scissor mechanism (green). A passive spring mechanism (bottom left) is used to balance the weight of the arm. With joint angles $\theta_1,\theta_2$, $\theta_3$ and motor torques $\tau_1$, $\tau_2$. $\tau_3$. Adopted from \cite{Ilg2025}}
\label{fig:deviation}
\end{figure}

\subsubsection{Mechanical Architecture and Actuation Concept}
The assistive device provides three actively actuated DoFs consisting of two rotational joints in the horizontal plane (x, y) and one vertical prismatic joint (z). To minimize moving mass and apparent inertia, all actuators are mounted on the fixed base of the system and connected to the joints through remote transmission mechanisms.
A ball-spline mechanism transmits torque through the vertically translating stage while allowing unrestricted motion along the z-axis, see Fig.\,\ref{fig:deviation}. The distal rotational joint is actuated via a cable-driven Capstan transmission integrated into the scissor mechanism. Routing the cables through the scissor linkage maintains nearly constant cable lengths throughout the workspace. This enables bi-articular actuation of the elbow joint using proximally mounted actuators, thereby avoiding additional moving masses on the robotic arm. A passive spring-based gravity compensation mechanism further reduces user-perceived loads and improves transparency. The user is connected through a forearm interface incorporating three passive rotational DoFs, which enables natural arm motions while preventing parasitic interaction moments \cite{Ilg2025}.

\subsubsection{Kinematic Model}
\label{sec:kin}
The device provides three actuated Cartesian DoFs: two coplanar links of equal length $L_1=L_2=L=0.45$\,m position the end-effector in the horizontal plane, while a third stage of effective length $L_z=0.30$\,m sets the vertical coordinate. Both planar joint angles $\theta_1$ and $\theta_2$ are measured in the device base frame; $\theta_3$ actuates the vertical stage.

\paragraph{Forward kinematics}
The end-effector position follows as
\begin{equation}
\begin{aligned}
x &= L\,(\cos \theta_1+\cos \theta_2),\\
y &= L\,(\sin \theta_1+\sin \theta_2),\\
z &= 2L_z\sin \theta_3 .
\end{aligned}
\label{eq:fk}
\end{equation}

\paragraph{Jacobian and decoupling}
Differentiating~\eqref{eq:fk} yields the end-effector Jacobian
\begin{equation}
J=\begin{bmatrix}
-L\sin \theta_1 & -L\sin \theta_2 & 0\\
 L\cos \theta_1 &  L\cos \theta_2 & 0\\
 0 & 0 & 2L_z\cos \theta_3
\end{bmatrix}.
\label{eq:jac}
\end{equation}
Its block-diagonal structure shows that the vertical axis is \emph{kinematically decoupled} from the horizontal plane: $z$ depends only on $q_3$, and its Jacobian column has no planar component. This decoupling is what permits the axis-wise control partition of Sec.~\ref{sec:control}---closed-loop force control on $z$ without cross-talk into the open-loop $x$--$y$ mapping.

\paragraph{Inverse kinematics}
Applying sum-to-product identities to the planar equations of~\eqref{eq:fk} gives a closed-form solution. With
$r=\sqrt{x^2+y^2}$, $\alpha=\operatorname{atan2}(y,x)$ and $\Delta=\arccos\!\big(r/2L\big)$,
\begin{equation}
\theta_1=\alpha-\Delta,\quad \theta_2=\alpha+\Delta,\quad
\theta_3=\arcsin\!\big(z/2L_z\big),
\label{eq:ik}
\end{equation}
where the sign of $\Delta$ selects the elbow branch; the physical build corresponds to $\theta_1<\theta_2$ (both referred to the base frame). Evaluating \eqref{eq:ik} at the three ADL operating points used throughout this work gives the joint configurations in Table~\ref{tab:ik}.

\begin{table*}[t]
\centering
\caption{Operating points, inverse-kinematic solution, and force-capability
analysis. The guaranteed omnidirectional planar force is the
shoulder-torque--limited lower bound $\min(a_1,a_2)$ and is
\emph{configuration-independent}: it holds over the entire reachable
workspace, not only at $P_1$--$P_3$. The pose-dependent quantities are the
vertical capability $f_{z,\max}$, the maximum planar force (parallelogram
diagonal), the anisotropy (max/guaranteed), and the Jacobian manipulability
$\det J_{xy}$.}
\label{tab:ik}
\begin{tabular}{lccccccccc}
\toprule
& & & & & \multicolumn{5}{c}{Force capability}\\
\cmidrule(lr){6-10}
Point & $(x,y,z)$ [cm] & $\theta_1$ & $\theta_2$ & $\theta_3$
      & $f_{z,\max}$ [N] & Guar.\ $\|f_{xy}\|$ [N] & Max $\|f_{xy}\|$ [N] & Anisotropy & $\det J_{xy}$\\
\midrule
$P_1$ grasp bottle & $(0,50,30)$   & $33.75^\circ$ & $146.25^\circ$ & $30.00^\circ$ & $10.5$ & $9.89$ & $18.2$ & $1.84$ & $0.187$\\
$P_2$ fill glass   & $(-10,60,25)$ & $52.00^\circ$ & $146.93^\circ$ & $24.62^\circ$ & $10.0$ & $9.89$ & $15.0$ & $1.51$ & $0.202$\\
$P_3$ to mouth     & $(-10,30,30)$ & $39.00^\circ$ & $177.86^\circ$ & $30.00^\circ$ & $10.5$ & $9.89$ & $28.8$ & $2.91$ & $0.133$\\
\bottomrule
\end{tabular}
\end{table*}

\subsection{Transparency-Oriented Control Concept}
\label{sec:control}

The primary control objective is transparency: with zero commanded assistive force, the device must not impede voluntary motion, while any assistance is superimposed as a controlled Cartesian force at the end-effector. Following the maxim \emph{open-loop where possible, closed-loop where necessary}, the axis-wise control strategy is derived from the disturbance budget characterized on the prototype in~\cite{Ilg2025}. In the horizontal plane, the device is backdrivable and the torque--force relation is near-linear and current-inferable. Horizontal forces are therefore delivered open-loop via the Jacobian-transpose map without a dedicated planar sensor. 
Preliminary neuromechanical investigations within the iAssistADL project indicated that corrective forces of approximately 5\,N are sufficient to compensate relevant pathological movement components during representative ADLs~\cite{Ilg2025}. Along the vertical axis, however, the ball-spline transmission introduces a dry-friction load of approximately 3\,N~\cite{Ilg2025}, which is of the same order as the desired assistive force. Consequently, this axis is operated with closed-loop force control using a single uniaxial load cell. This partition also minimizes sensing hardware, consistent with the single-force-sensor concept proposed in~\cite{Ilg2025}.

\subsubsection{Open-Loop Control in the Horizontal Plane}
The desired planar force $\mathbf F^\star=[F_x^\star,F_y^\star,0]^\top$ is mapped to motor torques by
\begin{equation}
\boldsymbol\tau_{\text{motor}} = S\,G\,J^{\!\top}\mathbf F^\star,
\qquad G=\mathrm{diag}(N_i\eta_i),
\label{eq:torquemap}
\end{equation}
with $J$ and gear ratios as in Sec.~\ref{sec:kin}. The deployed code uses $\eta_i=1$, so mechanical losses appear as a systematic under-delivery by the factor $\eta_{1,2}\in[0.92,0.96]$. The efficiencies were estimated experimentally from static force-transmission measurements by comparing commanded motor torques and measured end-effector forces. This is exactly the gap between the code-nominal command and the measured-efficiency capability envelope of Sec.~\ref{sec:kin}, which is thus the conservative, physically realized bound. The accuracy of the $J^{\!\top}$ map is not verified against a reference joint torque but end-to-end through the force-tracking results of Sec.~\ref{sec:results}.

\subsubsection{Gravity and Friction Compensation in z}
The $z$-axis uses a dual approach consisting of feedforward compensation for predictable disturbances and feedback control for residual force errors. The feedforward component compensates gravity and friction effects that would otherwise impair transparency. The quasi-static gravitational load is offloaded mechanically by the passive counterbalance, see Fig.\,\ref{fig:deviation}. No gravity term therefore appears in the control law, preserving the actuator torque headroom reported in Sec.~\ref{sec:kin}.

\subsubsection{Closed-Loop Force Control in the Vertical Direction}
Unlike the horizontal axes, accurate force rendering on the $z$-axis cannot be achieved using feedforward control alone because the ball-spline transmission introduces substantial friction and stick--slip effects. Therefore, the residual force error is regulated by a PI controller based on the load-cell measurement,
\begin{equation}
u_z = F_z^\star + C(s)\,\big(F_z^\star-\hat F_z\big),
\label{eq:zlaw}
\end{equation}
which is realized as motor torque through the $z$-row of~\eqref{eq:torquemap}, $\tau_{\text{motor},z}=N_z\,(2L_z\cos \theta_3)\,u_z$, with $\eta_z=1$ in code. The true loss $\eta_z=0.75$ enters as a multiplicative plant gain that the integral action rejects, so the vertical axis is not subject to the under-delivery of the horizontal plane. The nominal operating point is $F_z^\star=0$: the device renders itself force-neutral by canceling friction and driving the residual interaction force to zero, and assistance is obtained by superimposing $F_z^\star\neq 0$ tracked by the same regulator. The controller was tuned to minimize perceived resistance while maintaining low steady-state force errors during user interaction.

\subsubsection{Force Controller Design}
A PI compensator is used,
\begin{equation}
C(s)=K_p+\frac{K_i}{s},
\label{eq:pi}
\end{equation}
the derivative term being omitted to avoid amplifying load-cell noise in the tremor band. The loop runs at $f_s=1$\,kHz ($T_s=1$\,ms).

The gains $K_p=0.1$ and $K_i=0.2$ were set by manual tuning on the identified $z$-axis plant, guided directly by the transparency objective: gains were increased until the perceived friction was suppressed and the device followed the user's motion freely, but kept below the level at which the loop began to overcorrect and inject its own resistance. This subjective, interaction-oriented criterion is deliberate---since the human user is the ultimate arbiter of transparency, a felt-behavior tuning is more representative of the design goal than a model-based rule optimized for reference tracking. In our tests, it yielded smoother, less intrusive behavior than a Ziegler--Nichols tuning of the same loop, which produced an over-aggressive, oscillatory response inconsistent with the transparency requirement.

\subsection{Experimental Validation}
\label{sec:results}
The validation targets the two claims of the control concept (Sec.~\ref{sec:control}): that the horizontal plane is transparent enough for open-loop operation, and that the closed-loop $z$-axis renders forces accurately across the ADL workspace. Accordingly, the study is in two parts. First, the per-axis mechanical transparency is characterized by friction measurements; these results are the empirical basis for the open/closed control partition. Second, the closed-loop force-rendering performance is assessed on a dedicated testbed under static and dynamic loading, quantifying force accuracy, tracking bandwidth, and practical operating limits.
\begin{figure} [thbp]
\centering 
\def\svgwidth{0.8\columnwidth}
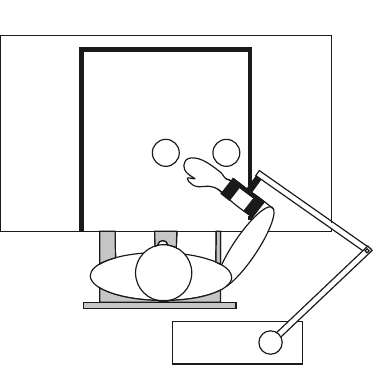
\caption{Test positions with glass (G) and bottle (B). F = Front, B = 
Bottom, L = Left, R = Right.}
\label{fig:testpoints}
\end{figure}

\subsubsection{Transparency Evaluation}
The transparency characteristics of the prototype were previously evaluated using a dedicated characterization test bench and are summarized here for completeness \cite{Ilg2025}. 
Each DoF was assessed individually under motion representative of the target ADLs. The rotational axes exhibited resistive torques below $0.1$\,N\,m, confirming highly transparent behavior in the horizontal plane. In contrast, the vertical axis showed a static friction force of approximately $3$\,N, dominated by the ball-spline transmission. This asymmetry motivated the hybrid control concept described in Sec.~\ref{sec:control}, combining open-loop force delivery in the horizontal plane with dedicated force-feedback control on the $z$-axis.

\subsubsection{Force-Rendering Evaluation}
Force-rendering experiments were run with the control algorithms executed in real time on a dSPACE MicroLabBox. The end-effector was rigidly coupled to a six-axis force/torque sensor (ATI Axia80, Novanta Inc., USA) on a fixed support structure. The ATI sensor served solely as an independent ground-truth reference and was not part of the control loop. All force commands were generated by the controller and compared against this external measurement. Experiments were repeated at the three ADL operating points of Table~\ref{tab:ik} (P1 grasp, P2 pour, P3 drink), illustrated in Fig.~\ref{fig:testpoints}.

Step responses were evaluated by the rise time $t_r=t_{90\%}-t_{10\%}$ and the steady-state force error $e_{ss}=F_{\mathrm{ref}}-F_{\mathrm{meas}}$. Overall tracking accuracy is reported as the root-mean-square error
\begin{equation}
\mathrm{RMSE}=\sqrt{\tfrac{1}{N}\textstyle\sum_{i=1}^{N}
\big(F_{\mathrm{ref},i}-F_{\mathrm{meas},i}\big)^2}.
\label{eq:rmse}
\end{equation}
For sinusoidal excitation, the frequency response is characterized by the amplitude ratio $A_r=A_{\mathrm{meas}}/A_{\mathrm{ref}}$ and the phase lag $\phi=360^\circ\,\Delta t/T$, with $\Delta t$ the reference-to-measurement delay and $T$ the excitation period. The usable frequency range is defined as the interval over which $A_r$ and $\phi$ remain within acceptable bounds without the onset of sustained oscillation.

At each operating point, force commands were applied along the individual Cartesian axes $x$, $y$, and $z$, complemented by a diagonal command $\mathbf F_{\mathrm{diag}}=[2,\,2,\,0]^\top$\,N to probe planar superposition. Four excitation types were used:
\begin{itemize}
\item \emph{Step responses} of $2$\,N and $5$\,N, characterizing rise time,
      settling, steady-state error, and stick--slip.
\item \emph{Sinusoidal tracking at $2$\,Hz} (amplitudes $2$\,N and $5$\,N),
      quantifying tracking accuracy and phase lag under moderate dynamics.
\item \emph{Sinusoidal tracking at $4$\,Hz} (same amplitudes), probing the
      dynamic limit within the tremor band and assessing suitability for
      future tremor compensation.
\item \emph{Frequency sweeps} at constant $2$\,N amplitude, from $0.05$\,Hz
      to $5$\,Hz over $100$\,s, yielding the amplitude/phase characteristic
      and the practical operating limit; runs were terminated at the onset
      of visible sustained oscillation.
\end{itemize}

\section{Results}
\label{sec:results-data}

\subsection{Force Capability}
The measured-efficiency capability analysis yields a \emph{guaranteed omnidirectional} planar force of $9.89$~N that is \emph{configuration-independent}: it is fixed by the shoulder-joint torque limit through the tighter of the two half-plane (slab) constraints, $\min(a_1,a_2)$ with $a_1=9.89$~N and $a_2=10.33$~N, and, therefore, holds over the \emph{entire reachable workspace} rather than only at the three ADL poses. The remaining, pose-dependent capabilities are summarized in Table~\ref{tab:ik}: the vertical axis provides $f_{z,\max}=10.0$--$10.5$~N, while the \emph{maximum} attainable planar force (the parallelogram diagonal of the two joint-torque constraints) ranges from $15.0$~N at $P_2$ to $28.8$~N at $P_3$. The force anisotropy---the ratio of this maximum to the guaranteed omnidirectional force---grows from $1.51$ to $2.91$ as the manipulability $\det J_{xy}$ degrades toward $P_3$. Since the entire $2$--$5$~N test range lies well inside the conservative
\emph{guaranteed} envelope at every reachable configuration, the experiments below probe rendering fidelity, not saturation.

\subsection{Transparency}
The transparency characteristics previously reported in \cite{Ilg2025} confirm the design assumptions underlying the proposed control architecture. The transparency evaluation demonstrated low mechanical resistance in the horizontal degrees of freedom, with average resistive torques below 0.1\,Nm for both rotational joints. In contrast, the vertical prismatic axis exhibited higher resistance due to the ball-spline mechanism. Despite this limitation, the device reduced the reachable workspace by only 2.2\,\%, indicating minimal interference with movements required for ADLs. Subjective evaluations with participants further indicated that horizontal motions were perceived as highly transparent, while slight resistance was noticed during vertical movements. These findings motivated the use of open-loop force rendering in the horizontal plane and dedicated force feedback on the vertical axis \cite{Ilg2025}. 

\subsection{Static Force Rendering (Step Response)}
Table~\ref{tab:step} summarizes the step-response behavior. In the horizontal plane, the open-loop map reaches the commanded level within $\approx 16$\,ms and settles in $\approx 1.3$\,s at a steady-state error of $2$--$5\%$, confirming that the $J^{\!\top}$ mapping renders planar force accurately without feedback. The residual amplitude ratio at low frequency (below) is consistent with the modeled efficiency loss $\eta_{1,2}\approx 0.92$--$0.96$, providing an independent cross-check of the capability model. The vertical closed-loop axis attains a comparable mean steady-state accuracy but with a markedly slower and far more variable transient (rise time $3.0$\,s, settling $7.0$\,s, $\sigma_{e_{ss}}\approx 3$\,N), consistent with stick--slip in the ball-spline: the friction feedforward removes the mean offset while residual stiction produces creep and run-to-run scatter.

\begin{table}[t]
\centering
\caption{Step-response metrics (mean over all cases).}
\label{tab:step}
\begin{tabular}{lccc}
\toprule
Metric & $F_x$ & $F_y$ & $F_z$\\
\midrule
Rise time $t_r$ (10--90\,\%) & $16$\,ms & $15$\,ms & $3.04$\,s\\
Overshoot & $33\%$ & $33\%$ & $28\%$\\
Settling time & $1.35$\,s & $1.24$\,s & $6.95$\,s\\
Steady-state error & $-0.04$\,N & $0.08$\,N & $-0.12$\,N\\
\quad (relative) & $2.1\%$ & $5.5\%$ & $-16\%$\\
\bottomrule
\end{tabular}
\end{table}

\subsection{Dynamic Force Tracking}
Sinusoidal results at $2$ and $4$\,Hz are given in Table~\ref{tab:sine}. The system exhibits an approximately constant transport delay of $\approx 0.13$--$0.19$\,s across both frequencies; because this delay is frequency-independent, it maps to a phase lag that grows roughly linearly with frequency ($-102^\circ$ to $-138^\circ$ at $2$\,Hz, $-179^\circ$ to $-253^\circ$ at $4$\,Hz). This delay dominates the raw RMSE ($3$--$5$\,N, comparable to the signal amplitude); the \emph{phase-compensated} RMSE is therefore additionally reported to isolate amplitude- and shape-tracking errors from the temporal delay.

Additional analysis was conducted to localize the origin of this transport delay. Comparison of motor-side encoder signals and end-effector force measurements revealed an average delay of approximately 98 ms between motor motion and force generation at the end-effector. This indicates that a substantial portion of the overall delay originates from the mechanical transmission path rather than from the digital control implementation.

In the horizontal plane, amplitude tracking is preserved up to $4$\,Hz (amplitude ratio $0.91$--$0.97$, phase-compensated RMSE $0.36$--$0.68$\,N), confirming that the open-loop map is dynamically well-behaved across the tested band. The vertical closed loop tracks accurately at $2$\,Hz ($A_r=0.98$) but exhibits a pronounced resonant amplification at $4$\,Hz ($A_r=1.87$, phase-compensated RMSE $3.4$\,N): the feedback loop \emph{overcorrects} at higher frequencies, injecting energy rather than tracking. This directly bounds the usable closed-loop bandwidth.

\begin{table}[t]
\centering
\caption{Sinusoidal tracking (mean): amplitude ratio $A_r$, transport
delay $\Delta t$, phase-compensated RMSE.}
\label{tab:sine}
\begin{tabular}{llccc}
\toprule
$f$ & Metric & $F_x$ & $F_y$ & $F_z$\\
\midrule
\multirow{3}{*}{$2$\,Hz}
 & $A_r$              & $0.91$ & $0.92$ & $0.98$\\
 & $\Delta t$ (ms)    & $142$  & $155$  & $191$\\
 & RMSE$_{\phi}$ (N)  & $0.36$ & $0.40$ & $1.24$\\
\midrule
\multirow{3}{*}{$4$\,Hz}
 & $A_r$              & $0.95$ & $0.97$ & $1.87$\\
 & $\Delta t$ (ms)    & $127$  & $124$  & $176$\\
 & RMSE$_{\phi}$ (N)  & $0.68$ & $0.57$ & $3.37$\\
\bottomrule
\end{tabular}
\end{table}

\subsection{Frequency-Sweep and Operating Limits}
The $0.05$--$5$\,Hz sweeps ($N=6$ per axis) confirm the picture above. The horizontal axes maintain high tracking quality (mean coherence $0.97$/$0.95$ for $x$/$y$) and near-unity mean magnitude ratio ($1.01$/$0.96$) across the band. The vertical axis degrades progressively: its mean magnitude ratio rises to $1.40$ with a peak gain of $2.7$ near $3.2$\,Hz, marking the resonant peaking that terminated the sweeps at the onset of visible oscillation. Mean planar direction errors of $9$--$13^\circ$ reflect the finite conditioning of the Jacobian map, consistent with the anisotropy reported for the poses (worst at P3, $2.91$).

Taken together, the horizontal open-loop channel is transparent and accurate across the full tested band, whereas the vertical closed-loop channel is accurate for quasi-static and low-frequency ($\lesssim 2$\,Hz) rendering but resonates beyond it. The device therefore meets the assistive (ADL) force-rendering target but does not, in its present tuning, satisfy the upper tremor band ($>2$\,Hz) on the $z$-axis.

\begin{figure}[t]
  \centering
  \setlength{\figurewidth}{0.85\columnwidth}
  \setlength{\figureheight}{0.62\figurewidth} \includegraphics{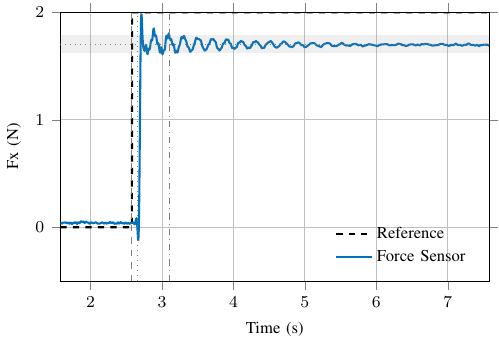}
  \caption{Open-loop step response of the $F_x$ axis. Reference (dashed) versus measured force (solid); shaded region is the $\pm5\%$ settling band.}
  \label{fig:step_fx}
\end{figure}

\begin{figure}[t]
  \centering
  \setlength{\figurewidth}{0.85\columnwidth}
  \setlength{\figureheight}{0.62\figurewidth}   \includegraphics{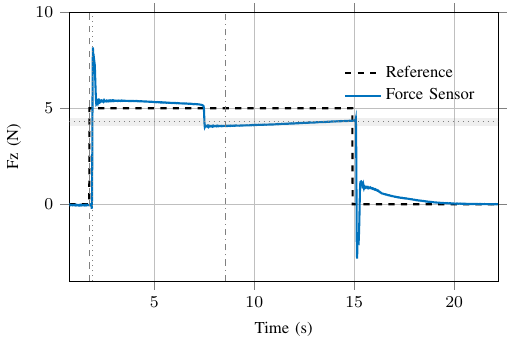}
  \caption{Stick--slip limit cycle on the vertical axis for the $F_z=5$\,N step reference at pose P3. The commanded control torque (top) ramps under integral action while the ball-spline carriage is stuck, breaks away and overshoots on slip, then re-sticks at a new offset, producing a sustained oscillation of the measured force (bottom) about the reference (dashed). The series compliance of the lightweight distal links stores and discharges elastic energy on each slip event, so the integrator hunts across the stiction dead-band without converging to a unique equilibrium.}
  \label{fig:limitcycle}
\end{figure}

\begin{figure}[t]
  \centering
  \setlength{\figurewidth}{0.85\columnwidth}
  \setlength{\figureheight}{0.62\figurewidth}   \includegraphics{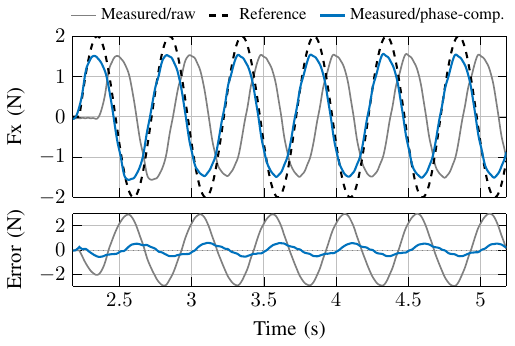}
\caption{Sinusoidal force tracking on the horizontal $F_x$ axis. Top: the commanded reference (dashed), the raw measured force (grey), and the phase-compensated response (blue). The raw signal reproduces the reference amplitude but is shifted by the near-constant transport delay ($\approx140$\,ms), which at this frequency manifests as a large phase lag; removing this delay recovers the phase-compensated trace, which closely overlays the reference. Bottom: instantaneous tracking error before (grey) and after (blue) phase compensation, showing that the raw error is dominated by the pure time delay---its removal collapses the error to the residual amplitude- and shape-tracking component (phase-compensated RMSE~$=0.36$\,N at $2$\,Hz), showing that the raw RMSE is dominated by the measured transport delay.}
\label{fig:fx-sine}
\end{figure}

\section{Discussion}
\label{sec:discussion}

The presented results demonstrate that the primary design objective of the iAssistADL platform, namely the combination of high mechanical transparency and effective force generation, was largely achieved. This trade-off has been identified as a central challenge in assistive upper-limb robotics, where supportiveness and transparency often represent conflicting design goals \cite{Arnold2025}. Furthermore, the system provided a guaranteed omnidirectional force capability of approximately 10\,N and demonstrated accurate force rendering in the horizontal plane across the tested frequency range.
The obtained results suggest that the platform is particularly suitable for supporting table-based ADLs such as drinking, pouring, and object manipulation, which were the primary target scenarios during the design process. These activities are frequently impaired by pathological movement components including dysmetric overshooting in cerebellar ataxia and action tremor during object transport. The high transparency observed in the horizontal plane is especially relevant for these scenarios, as the dominant movement components occur parallel to the table surface. For ataxia-related movement deficits, corrective forces are primarily required at relatively low frequencies and during goal-directed reaching movements. This interpretation is consistent with findings by Bhanpuri et al., who demonstrated that subject-specific external force compensation can improve reaching accuracy in patients with cerebellar ataxia \cite{Bhanpuri2014,Grow_Bastian_Okamura_2014}. The measured force capability and dynamic performance therefore indicate that the presented platform already provides a suitable basis for future compensation of such impairments. Furthermore, the demonstrated horizontal force-rendering performance suggests that compensation of low-frequency tremor components may also be feasible. Recent biomechanical and robotic investigations have similarly indicated that interaction forces in the order of magnitude generated by the present platform are sufficient for clinically relevant upper-limb tremor suppression \cite{ShaganShomron2025}.
The main limitation identified during the experiments was the friction of the ball-spline mechanism used for the vertical axis. While the horizontal degrees of freedom exhibited highly transparent behavior, the ball spline introduced static friction and stick-slip effects that limited the achievable closed-loop performance. This behavior resulted in increased settling times during step responses and resonant amplification at frequencies above approximately 2\,Hz. The results further suggest that the observed behavior arises not only from friction itself but from its interaction with the compliance of the lightweight arm structure. Consequently, improving the vertical-axis dynamics will require both refined friction compensation and further optimization of the stiffness-to-mass ratio of the moving structure.
A further limitation is the approximately 140\,ms transport delay observed during dynamic force-rendering experiments. Additional investigations revealed that roughly 98\,ms of this delay originate from the mechanical transmission path between motor actuation and force generation at the end-effector, indicating that the dominant contribution is mechanical rather than computational. While this delay can be compensated for offline when characterizing force-rendering performance, it would directly affect future real-time tremor suppression strategies and therefore represents an important area for further investigation.

\subsection{Limitations}
\label{sec:limitations}

Several limitations should be considered when interpreting the presented results. First, all force-rendering experiments were conducted on a rigid, instrumented testbed rather than in direct interaction with a human user. Consequently, the measured dynamics may differ under realistic interaction conditions, where the mechanical impedance of the human arm could alter the observed stick--slip behavior and closed-loop response. Furthermore, transparency and force-rendering performance were evaluated only at three representative ADL configurations, such that broader multi-axis interactions and workspace-dependent effects remain to be investigated. Second, the dynamic evaluation was limited to excitation frequencies up to 5\,Hz and to a set of representative operating conditions. While this range covers the frequency band relevant for many ADL-related assistive scenarios and low-frequency ataxic movement deviations, conclusions regarding higher-frequency behavior require further experimental validation. In addition, the number of trials per condition was limited, and the vertical axis exhibited considerable variability, leading to correspondingly wider confidence intervals for the reported $z$-axis performance metrics.

\section{Conclusion}
\label{sec:conclusion}

This paper presented the iAssistADL platform, a transparency-oriented 3-DoF assistive device for future tremor and ataxia compensation during activities of daily living. By combining a lightweight mechanical architecture with a hybrid force-control strategy, the system achieved low mechanical resistance during voluntary motion while maintaining a force capability of approximately 10\,N. Experimental results demonstrated highly transparent behavior in the horizontal plane and accurate force rendering across representative ADL configurations. The main limitation was identified as friction-induced stick-slip behavior of the vertical axis, which currently restricts the achievable bandwidth in this direction. Overall, the presented results establish the iAssistADL platform as a promising basis for future compensation of pathological upper-limb movement components. The next step will be the evaluation of the system in dynamic human-in-the-loop experiments and patient studies.

\section*{Ethics Statement} 
The experiments reported in this paper were conducted exclusively on a rigid laboratory testbed and did not involve human participants. Ethics approval (approval no. 2026-0144-BO) refers to previous and ongoing human-subject studies within the broader iAssistADL project. The transparency ratings cited from \cite{Ilg2025} were obtained under the corresponding approval procedures.

\addtolength{\textheight}{-11cm}   

\bibliographystyle{IEEEtran}
\bibliography{IEEEabrv,bibliography}

\end{document}